\documentclass[a4paper,10pt]{article}
\usepackage[utf8]{inputenc} 
\usepackage[T1]{fontenc}    
\usepackage[top=2cm,bottom=2cm,left=2cm,right=2cm]{geometry}
\usepackage{hyperref}       
\usepackage{url}            
\usepackage{booktabs}       
\usepackage{amsfonts}       
\usepackage{nicefrac}       
\usepackage{microtype}      
\usepackage{graphicx}
\usepackage{xcolor}         
\usepackage{glossaries}
\usepackage{float}
\usepackage{caption}
\usepackage{subcaption}
\usepackage{hyperref}
\usepackage{amsmath}
\usepackage{amssymb}
\newacronym{gnn}{GNN}{Graph Neural Network}
\newacronym{gcn}{GCN}{Graph Convolutional Network}
\newacronym{gat}{GAT}{Graph Attention Network}
\newacronym{xai}{XAI}{Explainable	Artificial Intelligence}
\newacronym{cnn}{CNN}{Convolutional Neural Network}
\newcommand{\V}{V}

\newcommand{\X}{\textbf{X}}
\newcommand{\A}{\textbf{A}}

\newcommand{\C}{\mathcal{C}}

\title{EiX-GNN: Concept-level eigencentrality explainer for graph neural networks}

\author{%
	Adrien Raison, Pascal Bourdon, David Helbert\\
	University of Poitiers\\
	Poitiers, France \\
	\texttt{\{firstname.lastname\}@univ-poitiers.fr}
}

\begin{document}

\maketitle

\begin{abstract}

	Nowadays, deep prediction models, especially graph neural networks, have a major place in critical applications.
	In such context, those models need to be highly interpretable or being
	explainable by humans, and at the societal scope, this understanding may also be feasible for humans that do not have a strong prior knowledge in
	models and contexts that need to be explained. In the literature, explaining is a human knowledge transfer process regarding a phenomenon between an explainer and an explainee.
	We propose EiX-GNN (Eigencentrality eXplainer for Graph Neural Networks) a
	new powerful method for explaining graph neural networks that encodes
	computationally this social explainer-to-explainee dependence underlying in
	the explanation process.
	To handle this dependency, we introduce the notion of explainee concept assimibility which allows
	explainer to adapt its explanation to explainee background or expectation.
	We lead a qualitative study to illustrate our explainee concept assimibility
	notion on real-world data as well as a qualitative study that compares, according to
	objective metrics established in the literature, fairness and compactness of our method with respect to performing state-of-the-art
	methods. It turns out that our method achieves strong results in both aspects.
\end{abstract}

\section{Introduction}
Graphs are widely used data structures involved in many real-world problems. \gls{gnn} \cite{scarselli_graph_2009} are artificial neural networks suited for such data structure.
For graph classification, node classification or link prediction tasks, \gls{gnn} models have shown impressive performances \cite{defferrard_convolutional_2016,zhang_link_2018}.
Regarding real-life deployment, \gls{gnn} models have shown impressive results for drugs design \cite{bapst_unveiling_2020}, web recommendations \cite{ying_graph_2018} or traffic forecasting \cite{derrow-pinion_eta_2021}.
A major drawback of those deep models is their occluded internal decisionnal processes, in particular in critical applications, it raises confidence, trustworthy, privacy and security concerns.
\gls{xai} is a set of methods that aims to tackle these issues by providing human-level meaningful insights about deep model internals by explaining how those models behave.
Explaining, understanding, or interpreting, although they are different notions, are intrinsically human dependent and context-dependent. So, it turns out that they are social notions.
One of those social requirements is that an explainer must adapt its explanation formulation according to the relative background of the explainee regarding the phenomenon to explain \cite{clough_global_2019,kim_interpretability_nodate}.
Several interesting \gls{xai} methods have been proposed for explaining graph neural network models but they often fail to take into account the social dependency when providing their explanations and rather focus only on the signal side of deep models to provide insights on deep model internals.
In this contribution we provide a social-aware explaining method that leverages background knowledge variability that is inherent in any social-related process while maintaining high score regarding state-of-the-art objective assessment metrics.
Firstly, we will frame the social context that the explanation process depends on. Then we will introduce our approach \footnote{Our implementation is available : \url{github.com/araison12/eixgnn}} in accordance with the numerical formulation of the social context. Then we provide the relevancy of our method against compared methods with a qualitative objective study on real-world applications and a quantitative objective study regarding objective metrics widely used in the literature.

\section{Related Work}
\gls{gnn} is firstly introduced by \cite{scarselli_graph_2009} with the message-passing scheme. They have been studied from the geometrical point of view and framed by \cite{bronstein_geometric_2021}, from this point of view, as the generalizing model of test-of-time models such as \gls{cnn} that have achieved main results in computer vision \cite{krizhevsky_imagenet_2012,russakovsky_imagenet_2015}
or Transformers \cite{vaswani_attention_nodate} in speech processing \cite{devlin_bert_2019} .
Genuine \gls{gnn} of \cite{scarselli_graph_2009} has been widely extended by \cite{satorras_en_2021,battaglia_relational_2018,defferrard_convolutional_2016,monti_dual-primal_2018,sanchez-lengeling_evaluating_2020,gilmer_neural_2017,gori_new_2005} including \gls{gcn} \cite{kipf_semi-supervised_2017} and \gls{gat} \cite{velickovic_graph_2018} that also achieve numerous results in their own fields.
Explaining methods are divided in several paradigms. The common approach is the model-based post-hoc local paradigm which consists in explaining each instance with including optimized deep model in the loop for furnishing explanations.
Under this paradigm, a serious amount of explaining methods to \gls{gnn} have emerged.
Attribution methods scope is to provide relevance to features regarding their impact on the classification, often under a white-box
approach \cite{battaglia_relational_2018,pope_explainability_2019} that have model internal insights either thank to model parameters and local behavior or
with relative contribution approach  \cite{yeh_fidelity_2019}.
Perturbation-based methods act in a black-box flavor (i.e., completely blind from model internal for
explaining) and their trouble model with node ablation procedure \cite{ying_gnnexplainer_2019}, or edges ablation \cite{luo_parameterized_2020,schlichtkrull_incorporating_2021} or counterfactual adjunctions \cite{lucic_cf-gnnexplainer_2022}. Heuristic search methods have also proven to be relevant for explaining as
well as generative model \cite{yuan_xgnn_2020}. Additionally, explaining methods suited for node classifiers has brought relevant results \cite{huang_graphlime_2022,vu_pgm-explainer_2020}.
Assessing the quality of those methods also remains a core challenge for the \gls{xai} community and some metrics have been proposed. They deal with explanation fidelity towards explained deep models. As well, sparsity measure is used to show the explanation compactness. These metrics are actually derived from an informal formulation of desirable property explaining methods have to fulfill \cite{jacovi_towards_2020,jacovi_aligning_2021,lipton_mythos_2018,murdoch_definitions_2019}.
All these approaches share a common assumption: explaining a deep classifier relies on finding relevant substructure on the instanced input that conserves the classifier behavior.
Besides aforementioned methods show interesting results, they always miss the inherently and intrinsically social dependence of the explanation process \cite{kim_interpretability_nodate,bas_scientific_1980,kulesza_too_2013,miller_explanation_2019}. Notably that an explainer has to share his knowledge regarding a phenomenon to an explainee in an explainee-understandable manner in order to have an effective explanation process profitable for the explainee \cite{bechtel_explanation_2005,glennan_rethinking_2002,chater_mental_2006,keil_explanation_2006}.
In this study, we first provide a relevant method that achieves stronger explanation results regarding state-of-the-art methods while fully encoding the explainee-knowledge dependency and more broadly the social context any explanation process in dependent on.


\section{Problem formulation}

Explaining helps human experts to inspect deep models, in order to show issues, blind spots or to prevent those models to potentially harm society.
For explaining machine learning problems, the social context \cite{selbst_fairness_2019} and human being factor \cite{noauthor_definition_nodate,miller_explanation_2019} are core elements.
Indeed, this explanation process involves an alignment of mental models. That alignment is between what the machine learning model is doing and
what the user thinks the model is doing. In order terms, to achieve this information trading it requires a set of arguments that conjointly machine and user are aware of and are able to deal with.
Explaining machine learning model is thus a human-centric process and in order to provide meaningful insight on how the model behaves to the user, the explanation process must be adapted \cite{kim_interpretability_nodate,clough_global_2019}.
Note that it is easier to shape machine outcomes representation than to force humans to think in far different way that they are used to think.
Consequently, it is the machine that has to be adapted to the user.
But when a user wants to have insightful explanations, they have to be expressed regarding a specific granularity. Indeed a user with a high level of knowledge has different explanation expectations than a user with less knowledge regarding the involved machine learning model.

More formally it can be reformulated as follows; explaining is a human knowledge transfer process involving an explainer (e.g., machine)
$E^{\star}$ and an explainee $\tilde{E}$ (e.g., user, engineer) concerning a phenomenon $P$ (e.g., machine learning model).
In order to have a profitable conversation (e.g., providing the explanation of $P$ from
$E^{\star}$ to $\tilde{E}$), both involved individuals must share a
common vocabulary set. It means that shared ideas must be expressed
upon a shared set of concepts by both individuals. This allows the
conversation to be profitable for them. For explanation purposes, the term profitable
means increasing the knowledge quantity of $P$ of $\tilde{E}$ thanks to $E^{\star}$
explanation.
For explaining, those concepts are framed as atomic parts that, when
carefully mixed, allow the explainer $E^{\star}$  to provide an explanation of $P$ to the
explainee $\tilde{E}$.
However, those elementary bricks are chosen conditionally to both knowledge
quantity of $E^{\star}$ and $\tilde{E}$ that are also dependent on $P$. Indeed if the
explainee $\tilde{E}$ has already a solid background or culture relatively to $P$ , basic
insights allowing shallow understanding of $P$ is already acquired by the
explainee $\tilde{E}$. Only finer details must be provided by the explainer $E^{\star}$ to explainee $\tilde{E}$ to have
total understanding of $P$. On the contrary, an explainee $\tilde{E}$ who has freshly begun to be
interested in $P$ must assimilate the coarsed concepts relative to $P$ before reaching
the finest ones with the explainer $E^{\star}$ having to adapt his vocabulary complexity in
order to be understandable.

%
%


\subsection{EiX-GNN}
\textbf{EiX-GNN} (\textbf{ei}gencentrality e\textbf{x}plainer for \textbf{g}raph \textbf{n}eural \textbf{n}etwork) is a post-hoc local model-based explaining method suited for any \gls{gnn} classifiers.
In our terminology, it provides its explanations according to a set of atomic concepts. These
concepts are for explanation processes what coins are for money exchanges, i.e., they are the elementary parts of the explanation process that explainers, when
explaining, will build their arguments upon those atomic concepts. Those concepts must be
carefully chosen by the explainer in order to match the explainee background on
the explained phenomenon. With assuming that the explainer has an optimal
knowledge of a phenomenon $P$ regardless concept selection, the concept
selection process depends on the background (relatively to $P$) of the explainee
and $P$. EiX-GNN has been designed to integrate this social dependence on the
explainee background given a phenomenon to explain.
Formally, we frame the set of explainee-admissible concepts as a probability
space $\mathcal{C}_p$ where concepts are $\C_p$-valued random variable. Parameter $p$ is the explainee concept assimibility
constrain. It is bounded as $p\in[0,1]$ and is proportional to the explainee
concept assimibility given $P$.
In the following, except for contrary mentions, we consider the phenomenon $P=(f^{\star},G,Y)$ where
$f^{\star}$ is an optimized \gls{gnn} classifier \footnote{see Appendix \ref{data:classifier} for details} of depth $D\in\mathbb{N}$ which has been trained on a
dataset $\mathcal{D}$ which $(G,Y)$ belongs to. In accordance with the graph formulation we have used before, we consider in the following a graph
$G=(\X,\A)$ has been composed of $N\in\mathbb{N}$ nodes and $M\in\mathbb{N}$ edges. We
also assume that the explainee has an explainee concept assimibility constraint
$p\in[0,1]$.
EiX-GNN provides its explanation based on a conditioned local and global
explainee-suited concept ordering. Firstly, we introduce the concept generation
procedure, then the global concept
ordering process which is the common thread of the overall explaining procedure is
described. Finally, the local concept ordering procedure is presented, this second step is a
refining procedure that highly precise at a node level the provided explanation.

\paragraph{Concept generation}
As mentioned above, concepts are atomic elements that allow the explainer
to provide its explanation. Given the explainee concept assimibility $p$,
concept $C_p$ is a $\C_p$-valued random variable. This variable is a
subgraph of $G$ such that $|C_p|=\lfloor |G|\times p\rfloor$. Our motivation from the signal
point of view is to describe an insightful subpart of the signal evolving on a subdomain of $G$.
Indeep in many deep-based data representation tiny but numerous low-level informations (e.g., high frequencies in the picture) are gathered along model depth to produce a unified high-level information (e.g., a probability distribution of classes that this picture belongs to). Classes probability is understandable by any person interested in deep learning approaches whereas high-frequency understanding is only doable for peoples with dedicated knowledge in image processing.
We have designed our concept generation process with respect to this data representation hierarchization, from detailed expert-understandable representation to commonly understandable representation, involves in deep representation methods.
Once determined the desired explanation granularity thanks to $p$, we need to sample those concepts from the initial graph $G$.
We have selected sampling approaches which depend either on a prior distribution or not. Sampling concept is thus a subgraph
sampling process which has a combinatorial aspect inherent of any subgraph
sampling problems. Concepts are key components of our approach, they have to be
carefully selected since they are providing our raw materials for conceiving
explanations. From all $\binom{|G|}{|C_p|}$ possible subgraphs we can derive
from $G$, some are more suited for providing explanation of $P$ than others.
Assuming a uniform relevance distribution for explaining $P$ among all those
subgraphs is not adapted, seamlessly, assuming that the sampling distribution is
$\mathbb{U}_{\binom{|G|}{|C_p|}}$ is not adapted either. We rather consider a
light importance sampling approach that quantifies the prior relevance
distribution of nodes
conditionally to $P$. For building such probability distribution, we apply a
node ablation approach that assesses the importance of nodes within their
neighborhood with respect to $P$. Formally, for a neighboring node
$v_j\in\mathcal{N}_i=\{v_j|(v_i,v_j)\in E\}\subset V$ of $v_i$. To quantify
node ablation importance we define a random variable $s: \V^2\rightarrow
	\mathbb{R^+}$ that measures the relative disturbance effect between two nodes
relatively to $P$ (e.g., relative $f^{\star}$
performance alteration impact of removing $v_j$ from $\mathcal{N}_i$). With assuming a uniform relevance
distribution $\mathbb{U}_{|\mathcal{N}_i|}$ \footnote{Note that we can extend our algorithm to node classification task. For explaining node $n$, we can consider to sample rather given $\mathbb{U}_{|\mathcal{N}_i\cap Hop(n)_L|}$ where $Hop(n)_D$ is the $D$-hop of node $n$.}, where of nodes composing $\mathcal{N}_i$, we defined the prior
relevance distribution $\alpha_P$ of the node $v_i$ conditionally to $P$ by:

\begin{equation}
	\alpha_P(v_i)=\underset{v_j\sim\mathbb{U}_{|\mathcal{N}_i|}}{\mathbb{E}}\left[s(v_i,v_j)|v_i,P\right]
\end{equation}

With a normalizing constant $F\in\mathbb{R^*}$ such that
$F^{-1}\sum_{v_i\in V}\alpha_{P}(v_i)=1$ we obtain a prior node importance probability distribution
that allows more efficient sampling process for determining pertinent
concepts with respect to $P$. Once such prior distribution is
determined, we sample in an i.i.d manner
$L\in\mathbb{N}$ realizations of $C_p$ which we denote by
$(C_i)_{i\in\{1,\dots,L\}}$ where each node composing the subgraph $C_i$ has
been sampled thanks to the prior node sampling distribution. Next, we will
present the procedure for hierarchizing those $L$ concepts relatively to $P$.

\paragraph{Global concept ordering}
Once concepts are sampled, we must find an ordering relationship in order to classify
their relevance according to $P$. Thanks to the prior node importance sampling
approach, we have already established such hierarchization but among all possible
subgraphs of $G$ with size $|C_p|$ which considerably reduces the research
perimeter of the optimal substructure that will explain $P$. Instead, here we
present an ordering method that hierarchies pair-wisely concepts among the $L$
sampled concepts.
Considering these $L$ concepts, we build an operational research tree with $G$
as root and these $L$ concepts as leaves. Without any further works, we do not
know yet if a concept $C_i$ is more relevant than another concept $C_j$
for explaining $P$. In order to provide such ordering, we derive from the
sample a complete graph $K_L$ where each node represents a concept and edge of
$K_L$ represents the relative similarity between two concepts relatively to $P$.
Since in this context graph are seen as signal evolving on a precise
deformation, we take into account
each both aspects for quantifying concept similarities pair wisely.

\paragraph{Relative concept domain similarity} We define the domain similarity
between two concepts $C_i,C_j\in(C_l)_{l\in\{1,\dots L\}}$ as the relative edge density between $C_i$ and $C_j$.
The graph edge density of a concept $C_i$, denote $d(C_i)$ is the ratio between the actual edges
composing $C_i$ over the total number of possible edges $C_i$ can be composed
of. For a graph $G=(\X,\A)$ with $N$ nodes and $M$ edges, it is defined as follows:
\begin{equation}
	d(G)=\frac{(2\times\boldsymbol{1}_{\{\A=\A^{T}\}}+\boldsymbol{1}_{\{\A\neq\A^{T}\}})M}{N(N-1)}
\end{equation}

It measures how $C_i$ tends to be a complete graph.
We choose this measure because of the local aggregation operation involve in
many \gls{gnn} models. We know that complete subgraphs aggregate much more signal than sparser ones.
This is due to the local invariance operation involved in any geometric deep learning models, especially \gls{gnn} \cite{bronstein_geometric_2021}.
Admittedly, aggregating numerous neighboring signal does not imply to aggregate more relevant information of this neighboring than with more sparse structures.
Nevertheless, doing so will produce in statistically a fairer estimation of the local information relevance given $P$ than it can be done on more degenerated localities.
In other terms, this approach allows yielding statistically more fidel local representations of $P$.
%

\paragraph{Relative concept signal similarity}
The concept signal similarity quantifies how similar $f^{\star}$ behaviors
are with respect to $P$ when the signal is propagated over a given concept subdomain and when it is propagated over another subdomain supplied by another concept.
Let assume that we considered two concepts $C_i$ and $C_j$, the case where $C_i$
is similar to $C_j$ given $P$  means that $f^{\star}$ sees equivalently $C_i$ and
$C_j$. Considering $C_i$ does not provide any added value than
solely considering $C_j$ itself, with respect to $P$. As a similarity
metric between two concepts $C_i$ and $C_j$ we use the Kullbach-Liebler
divergence of both inferred probability distributions of $C_i$ and $C_j$ thanks
to $f^{\star}$. Formally, we frame
$s_{f^{\star}}(C_i,C_j): \mathcal{C}_p\rightarrow \mathbb{R^+}$ as the $f^{\star}$
behavior similarity
metric concerning $C_i$ and $C_j$ by:
\begin{equation}
	s_{f^{\star}}(C_i,C_j)=\frac{1}{2}[D_{KL}(\textbf{f}^{\star}(C_i)||\textbf{f}^{\star}(C_j))+D_{KL}(\textbf{f}^{\star}(C_j)||\textbf{f}^{\star}(C_i))]
\end{equation}

where $D_{KL}(\cdot||\cdot)$ denotes the Kullbach-Liebler divergence.
This metric is widely used in machine learning problems. It has been deeply studied in various applications, especially for deep-based classification problems. In such problems, data representation is rendered as probability laws, and Kullbach-Liebler divergence is used in this context to quantify the similarity between inferred and groundtruth probability laws.

\paragraph{Domain and signal relevancy unification}
We introduce here our process for unifying those two modalities in order to have a global concept ordering.
For each modality, we have obtained real values quantifying the relative marginal relevance of each concept. Given a concept, we obtained the relative joint relevancy (i.e., the relative global concept relevancy) by multiplying each marginal value (i.e., the relative concept domain revelancy value and the relative concept signal similarity).
More formally , computing the relative global ordering concept revelancy between concept $C_i$ and $C_j$ is the real value $a_{i,j}$ defined by:
\begin{equation}
	a_{i,j}=\frac{d(C_i)}{d(C_j)}\times s_{f^{\star}}(C_i,C_j),\forall i,j\in\{1,\cdots,L\}^2
\end{equation}
From now, instead of considering $L\times L$ relative local ordering values, we want to hierarchies globally those $L$ concepts with a global concept ordering strategy.
Since the relative global concept relevancy can be seen as interactive strengths between concepts, a natural representation to render those relational interactions are graph themselves.
Thereby, we consider the graph $K_L$ composed of $L$ nodes that represent concepts with an adjacency matrix $\A(K_L)$
which entries are determined by $a_{i,j}$.
However, although we have obtained most $L$ meaningful concepts thank to previous processes, we look, at global scope, for the most dissimilar concepts pairwise ordonnance. Indeed, from explanation point of view, two highly similar concepts $C_i$ and $C_j$ (i.e., $a_{i,j}$ is low) bring similar insights regarding the explanation. In other terms, it produces redundant information that is unnecessary and may flood and alter the user understanding of phenomenon $P$.
Higher values in $\A(K_L)$ stand for those less redundant concepts (relatively) regarding the explanation of $P$.
But what is the concept that is both relevant and less redundant among concept candidates ? This question can be reformulated under graph theory by which node has the higher normalized centrality.
A good approach is to compute the PageRank \cite{page_pagerank_1999} of each node of $K_L$. Once obtained it gives a total ordering relation between nodes of $K_L$ (i.e., concepts).
Formally, we consider:

%
\begin{equation}
	\widehat{\A(K_L)}=\boldsymbol{\Lambda}(\A(K_L)\textbf{e})^{-1}\A(K_L)
\end{equation}
which is the normalized version of $\A(K_L)$ where $\textbf{e}$ the unit vector of
size $L$ and for any fixed-size vector $\textbf{x},\boldsymbol{\Lambda}(\textbf{x})$ denotes the diagonal matrix containing $\textbf{x}$ in its diagonal.

$\widehat{\A(K_L)}^{T}$ is a stochastic version of $\A(K_L)$ that by definition always admit a right eigenvector $\textbf{r}$ with eigenvalue equal to 1.
Under this context, this eigenvector $\textbf{r}$ defines a probability law and its components are \textit{PageRank} values of each node.
Regarding the explanation process, the \textit{PageRank} centrality measures yield a global concept ordering scheme where concept candidate with highest \textit{PageRank} value is the explicative representant that proposes the less information redundancy while being in both modalities relevant.
This global concept ordering procedure allows to tremendously shrink the search space to find relevant subgraphs which is known to be a combinatorial problem.
Once addressed this refinement, we can go further and we propose a less coarsening approach that assesses relevancy at $G$ nodes scope that we framed as the local concept ordering procedures.

\paragraph{Local concept ordering}
Considering only subgraph-level as the only set of explanation arguing terms may lead to incomplete formulation of explanations.
Indeed, although in underlying manner, nodes relevance is already partially encoded in concept relevance quantification processes, nodes composing these subgraphs may have themselves their own role on the global concept ordering outcomes; that given a concept; node has no-uniform contribution to this outcome.
Besides that purely signal-based argue, in many real-life applications, nodes may represent atoms for  molecule representations or city on a roadmap for traffic forecasting and therefore have their own semantic embeddings that may not be rendered in a single subgraph-level focus.
That is why, including such node-level data have to be included in our explanation conception pipeline.
To carefully quantifying the contribution of each node within a concept candidate $C_i$ we have exploited game theory.
It consists in computing the \textit{Shapley} value \cite{shapley_notes_1951} of each node $i$ composing $C_i$.
The Shapley value is a conceptual solution in cooperative game theory quantifying how important the marginal role of a player has in the game outcome.
Considering a coalition of $K\in\mathbb{N}$ players indexed within $Q=\{1,\dots,K\}$ playing a cooperative game with a game payoff $v: \mathcal{P}(Q)\rightarrow\mathbb{R}$ where $\mathcal{P}(Q)$ denotes all possible subsets of $Q$. The Shapley value of a player $i\in Q$, is defined by:
\begin{equation}
	\gamma_{Q}(i)=K\underset{j\sim\mathbb{U}_{K}}{\mathbb{E}}\left[\underset{S\subset Q\backslash\{i\}}{\mathbb{E}}\left[v(S\cup\{i\})-v(S))\hspace{0.1cm}|\hspace{0.1cm}|S|=j\right]\right]
\end{equation}
We denote further $\gamma_j(i)$ the \textit{Shapley} value of the node $i$ of concept $C_j$.

\paragraph{Global and local concept gathering}
Under our context, given a node $i$ that belongs to a concept $C_j$, computing the \textit{Shapley} value of $i$ required to consider all possible subgraphs of $C_j$ and compute, according to them, the perturbing effects of $i$ regarding $f^{\star}$ at $C_j$ scope.
Numerically, $\gamma_j(i)$ provides a precise concept relevance value of node $i$ belonging to $C_j$ regarding $P$ (\cite{oliver_graphsvx_2021,yuan_explainability_2021}). Note that this value $\gamma_j(i)$ remains dependent to $C_j$ definition.
From the computational point of view, the assessment requires $\mathcal{O}(2^{\lfloor |G|\times p \rfloor})$ inferences of $f^{\star}$ which can be intensive, even intractable in practice and is by definition dependent on the explainee concept assimibility constraint $p$.
To overcome this issue, we can estimate each $\gamma_j(i)$ by a Monte Carlo estimation strategy with an error rate bounded.
Those computations produce a set of $L$ node-level explaining assessments $(\boldsymbol{\gamma_j})_j\subset\mathbb{R}^{\lfloor |G|\times p \rfloor}$ where each $\boldsymbol{\gamma}_j$ is normalized by its $L_1$ norm. We then extend each $\boldsymbol{\gamma_j}$ to $\boldsymbol{\gamma_j^{ext}}\in\mathbb{R}^N$, such that for node $i$:
\[
	\boldsymbol{\gamma_j^{ext}}[i]=
	\begin{cases}
		 & \boldsymbol{\gamma_j}[i]\text{ if $i\in C_j$} \\
		 & 0\text{ otherwise}
	\end{cases}
\]
And we concatenate columns wisely each $\boldsymbol{\gamma_j^{ext}}$ defined for a explainee concept assimibility constrain $p$ in $\boldsymbol{\Gamma}_p\in\mathbb{R}^{N\times L}$.
Finally, our explanation map $\texttt{EiX-GNN}_{L,p}(P)$ \footnote{Code repository will be released after reviewing process.} of the phenomenon $P$ with an explainee concept assimibility constrain $p$ is algebraically defined as below:
\begin{equation}
	\texttt{EiX-GNN}_{L,p}(P)=\boldsymbol{\Gamma}_p\boldsymbol{\Lambda}(\textbf{r})\textbf{e}^T
\end{equation}

The explanation map $\texttt{EiX-GNN}_{L,p}(P)\in\mathbb{R}^N$ describes the relevance of each feature describing the phenomenon $P$ with respect to $p$. In the context of deep graph classification, it described the normalized relevance of each node composing $G$ regarding $f^{\star}$ with feature granularity of size $p$.
Now we lead a quantitative and a qualitative study on real-world applications as well as providing an impact study regarding the explainee concept assimibility that we have introduced.

%
%
\section{Results}
\subsection{Experimental setup}
\paragraph{Datasets}
To assess our method we have used four real-world datasets that are made of human intelligible features :  MNISTSuperpixels \cite{monti_geometric_2017}, PROTEINS \cite{borgwardt_protein_2005}, MSRC \cite{shotton_textonboost_2009,winn_object_2005}, REDDIT-BINARY \cite{yanardag_deep_2015}. These datasets are widely used in the literature for illustrating \gls{gnn} explainer. We give further details regarding those datasets in Appendix \ref{data:dataset}.
\paragraph{Learning procedures} We have used two main GNN configurations for classifying our instances. Either based on GCN or GAT modules, astonishingly both produce similar results in terms of test accuracy. But GCN-based model is less parametrized than GAT-based one, so we have selected GCN models.
Models architecture and learning setup are described in Appendix \ref{data:classifier}.
\paragraph{Comparing methods}
For comparing our results, we have retained three state-of-the-art methods that achieve strong results for explaining \gls{gnn}: GNNExplainer \cite{ying_gnnexplainer_2019}, SubgraphX \cite{yuan_explainability_2021}, PGExplainer \cite{luo_parameterized_2020}. We give further details regarding these methods in Appendix \ref{data:comparing} .
%
\paragraph{Objective assessment metrics}
Assessing explanation quality or relevance given a phenomenon often deals with requiring a $P$-specialist approval.
Context-free and objective method has been proposed for quantifying explanation method relevance. Two of them have been widely used in the literature \cite{oliver_graphsvx_2021,pope_explainability_2019,yuan_explainability_2021}, namely Infidelity \cite{yeh_fidelity_2019} and spatial sparsity \cite{oliver_graphsvx_2021,pope_explainability_2019,yuan_explainability_2021}. We have used them to lead our quantitative study.
Further details are provided in Appendix \ref{data:objmet}.

\subsection{Qualitative assessment: a real-world application}
For illustrating our method we have oriented our experiment in an omniscient setup: $L=70$ allowing drawning complex explanations with large argumentation basis and $p=0.05$ for focusing on finest data details.
We discuss afterward the marginal impact of each of these parameters in regard with the omniscient setup as a baseline.
Each instance of \textit{REDDIT-BINARY} is a discussion involving users with varying knowledge regarding the discussion topic.
Some users have a serious understanding of the subject and can be seen as experts. Explaining those discussions, in terms of user interactivity, consist in looking for those experts as mentioned in \cite{ying_gnnexplainer_2019}.
\begin{figure}[H]
	\centering
	\includegraphics[width=0.25\textwidth]{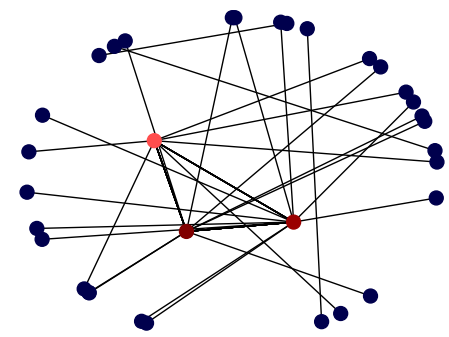}
	\includegraphics[width=0.25\textwidth]{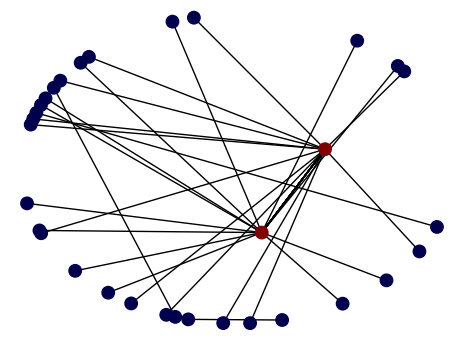}
	\includegraphics[width=0.25\textwidth]{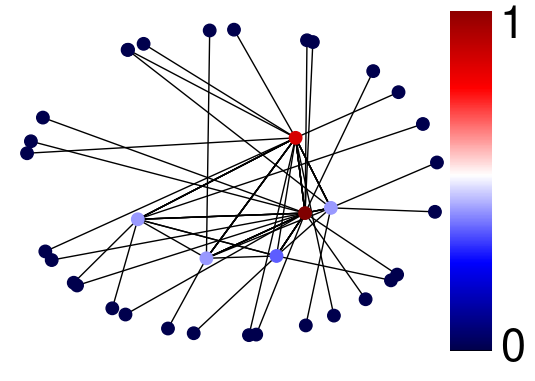}
	\caption{Threads explanation with EiX-GNN}
	\label{fig:res}
\end{figure}
It turns out that expert is actually users that are responding the most to all other users. In graph theory terminology, those experts are represented by node with highest relative degree.
Under the omniscient setup, EiX-GNN highlights those expert users and it locates them in a graph with low attribution to users with low interactivity (low knowledge) and high attribution to users with high interactivity (high knowledge), i.e., experts (Figure \ref{fig:res}). Those results are in accordance with those obtained in \cite{ying_gnnexplainer_2019}.
Now we measure the marginal impact of each of $p$ and $L$ on some thread explanations and we compared such explanations with the omniscient baseline which is seen as the practically upper bound of quality explanations that required both high understanding and high knowledge (i.e., retrieving only most relevant information, localizing carefully thread experts).
From the social point of view, we seek to qualify the marginal impact of these two parameters on the social expressiveness of EiX-GNN explanations. What we expect to get is to have uncomplete explanation with an uncomplete concept basis (i.e low $L$) and coarsed explanation with a large explainee concept assimibility constraint $p$, all in regard with a complete and finest explanation through the omniscient baseline.

\paragraph{Explainee concept assimibility constraint: qualitative impact}
\begin{figure}[H]
	\centering
	\begin{subfigure}{0.19\textwidth}
		\includegraphics[width=\textwidth]{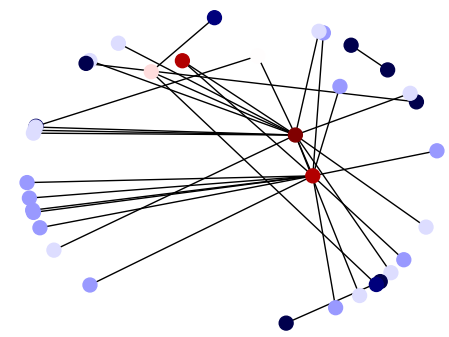}
		\caption{$p=0.7$}
		\label{fig:p7}
	\end{subfigure}
	\begin{subfigure}{0.19\textwidth}
		\includegraphics[width=\textwidth]{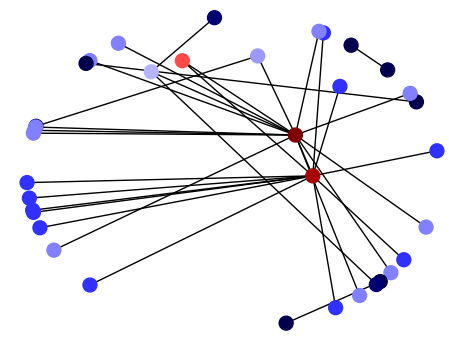}
		\caption{$p=0.3$}
		\label{fig:p3}
	\end{subfigure}
	\begin{subfigure}{0.19\textwidth}
		\includegraphics[width=\textwidth]{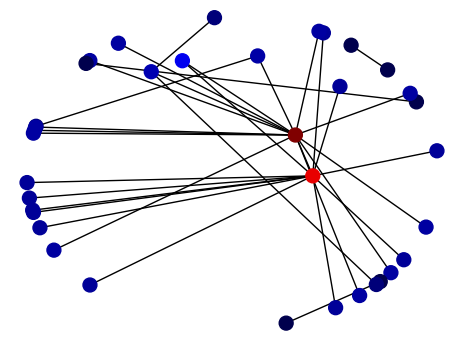}
		\caption{$p=0.2$}
		\label{fig:p2}
	\end{subfigure}
	\begin{subfigure}{0.19\textwidth}
		\includegraphics[width=\textwidth]{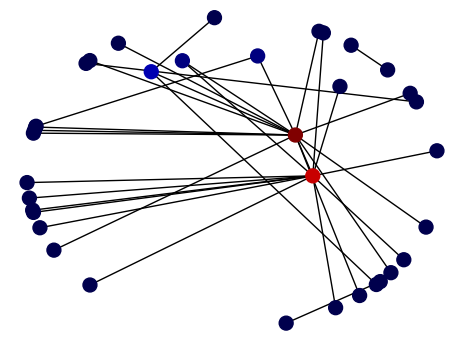}
		\caption{$p=0.1$}
		\label{fig:p1}
	\end{subfigure}
	\begin{subfigure}{0.19\textwidth}
		\includegraphics[width=\textwidth]{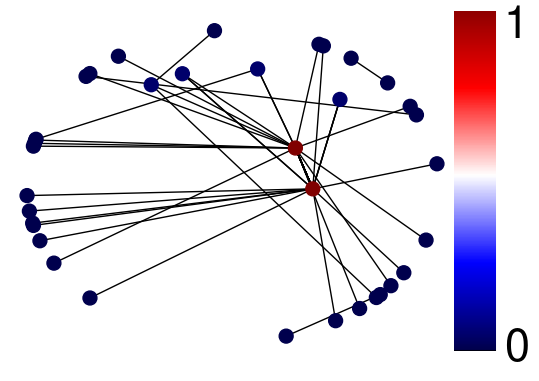}
		\caption{Omniscient baseline}
		\label{fig:po}
	\end{subfigure}
	\caption{Social expressiveness: explainee concept assimibility constraint variation}
\end{figure}
In this sequence of explanations of the same thread, we have made an explainee concept assimibility constraint variation and we have fixed the concept basis width.
Low value of $p$ stands for low explainee concepts assimibility constraint meaning that the explainee is able to reach finest understanding of the phenomenon.
Here, it signifies to be able to precisely recognize thread expert which is the most relevant information in the context of \textit{REDDIT-BINARY} classification.
The opposite scheme appears with high value explainee concept assimibility constraint.
We observe that as long as we raise the constraint penalty ($p$ decreases), we gain insights (Figures \ref{fig:p2}, \ref{fig:p1}) with respect to explanation precision and we incrementally increase the knowledge quantity until reaching the omniscient regime (Figure \ref{fig:po}).
\begin{itemize}
	\item $p=0.7$: explanation is based on large concepts providing coarse knowledge, single interactivity users have quite important role in the explanation and experts have higher explaining value. This explanation is not the finest one but allow explainee to have an imprecise but global view of the thread. For low knowledge requirements, this explanation is suitable (Figure \ref{fig:p7}).
	\item $p=0.3$: we observe here that specialist-level information is far more emphasized than previously. We have experts recognizing  and less insightful information are much discarded (single interaction users that are not specialists) (Figure \ref{fig:p3}).
	\item $p=0.2$: the previous tendency has been accelerated, specialist knowledge is far more mentioned than the poor knowledge (Figure \ref{fig:p2}).
	\item $p=0.1$: we have almost reached the omniscient regime and we have gained an understanding comparable to specialist one (Figure \ref{fig:p1}).
\end{itemize}
Globally we observe that this explainee concept assimibility constraint behaves as a social-aware explanation fine tuner. High constraint provides general-trended information, this information is general but imprecise. It provides a global idea about the underlying phenomenon. It thus allows non-specialist individuals to handle those explanation of the phenomenon.
As long as the constraint is raised, we tend to reach expert understanding of the explained phenomenon by including only finest details and discarding entities that only are able to supply generalities that are only dedicated to non-specialist peoples.
\paragraph{Number of concepts: qualitative impact}
\begin{figure}[H]
	\centering
	\begin{subfigure}{0.22\textwidth}
		\includegraphics[width=\textwidth]{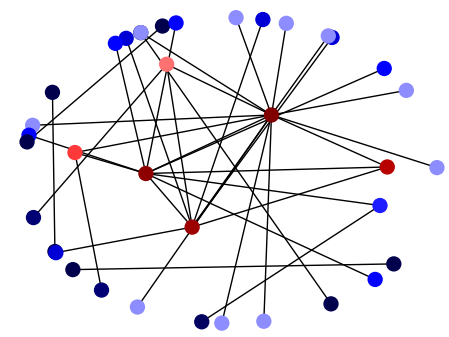}
		\caption{$L=10$}
		\label{fig:l10}
	\end{subfigure}
	\begin{subfigure}{0.22\textwidth}
		\includegraphics[width=\textwidth]{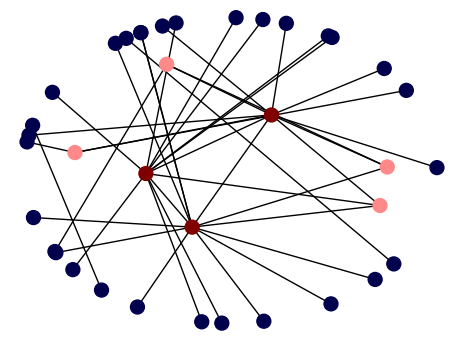}
		\caption{$L=30$}
		\label{fig:l30}
	\end{subfigure}
	\begin{subfigure}{0.22\textwidth}
		\includegraphics[width=\textwidth]{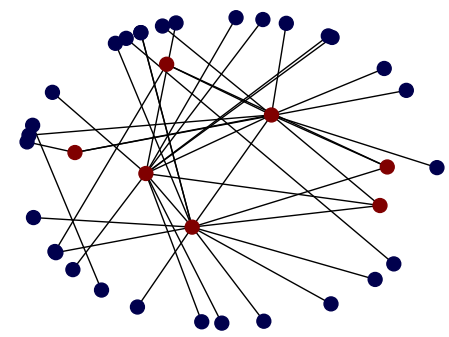}
		\caption{$L=50$}
		\label{fig:l50}
	\end{subfigure}
	\begin{subfigure}{0.22\textwidth}
		\includegraphics[width=\textwidth]{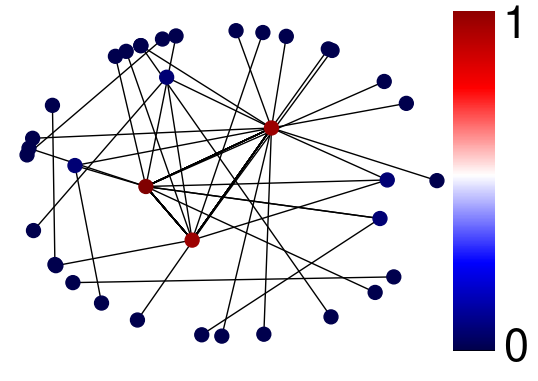}
		\caption{Omniscient baseline}
		\label{fig:lo}
	\end{subfigure}
	\caption{Social expressiveness: concept basis variation}
	\label{fig:lS}
\end{figure}
We observe that as long as the number of concepts $L$ increases from low width (Figure \ref{fig:l10}) to high-width (Figure \ref{fig:l50}) the explainer is able to furnish more and more precise explanations. So, as much as the concept basis width increases, we are getting closer to the omniscient baseline (Figure \ref{fig:lo}).
Actually, this behavior can be expected since if the explainer is able to provide explanation based on large arguments basis, we obtain precise and meaningful explanations \footnote{An analogous vision can be made with the neural network complexity that, if adapted to the learning task, allows to have powerful model.}.
\subsection{Quantitative assessment on real-world data}
\paragraph{Objectives metrics overall benchmarking}
As a global view regarding state-of-the-art methods, we have compared objective
metrics between each dataset and each method.
For \texttt{EiX-GNN}, we have used the omniscient setup presented above.
We find out that our method proposes numerically fewer infidel explanations with at least a factor $10^2$ on MNISTSuperpixel and REDDIT-BINARY and MSRC-21 and a factor 10 on PROTEINS and MSRC-9.
As well, our method outperforms other compared methods regarding the sparsity of explanation maps by at least a factor $10^3$ on MNISTSuperpixel and REDDIT-BINARY, a factor $10^4$ on MSRC-9, a factor $10^2$ on MSRC-21 and a factor 10 on PROTEINS.
We provide in Appendix \ref{data:objmet} a summarizing table (Table \ref{tab:1}) with a detailed version of these measurements.
\paragraph{Explainee concept assimibility constraint: quantitative impact}
\begin{figure}[H]
	\centering
	\begin{subfigure}{0.3\textwidth}
		\includegraphics[width=\textwidth]{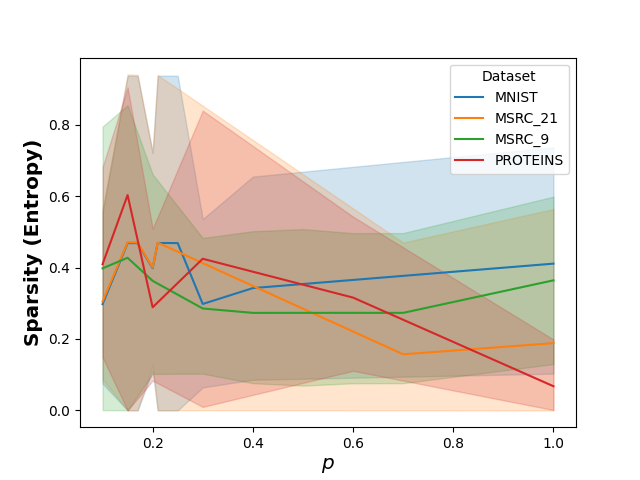}
		\caption{Von Neumann Entropy}
	\end{subfigure}
	\begin{subfigure}{0.3\textwidth}
		\includegraphics[width=\textwidth]{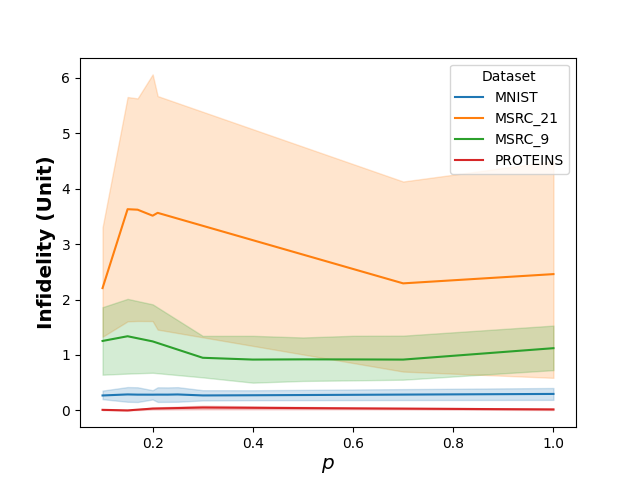}
		\caption{Infidelity (Unit)}
	\end{subfigure}
	\begin{subfigure}{0.3\textwidth}
		\includegraphics[width=\textwidth]{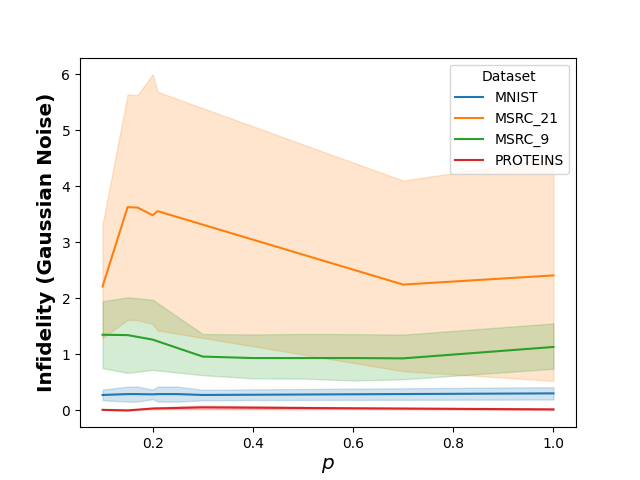}
		\caption{Infidelity (Gaussian noise)}
	\end{subfigure}
	\caption{Impact of $p$ regarding objective metrics}
	\label{p_bench}
\end{figure}
Regarding the explainee concept assimibility constraint, we find out that in average it does not have an impact on the infidelity of the explanation toward the classifier has shown in Figure \ref{p_bench}. Moreover, specialist-level explanation is more concise so inherently sparser as shown by Figures \ref{fig:p1}, \ref{fig:l50}.
It means that the value of $p$ does not impact the explanation quality provided by EiX-GNN and that EiX-GNN still provides relevant explanation regardless the explainee knowledge for a given phenomenon.
\paragraph{Number of concepts: quantitative impact}
\begin{figure}[H]
	\centering
	\begin{subfigure}{0.3\textwidth}
		\includegraphics[width=\textwidth]{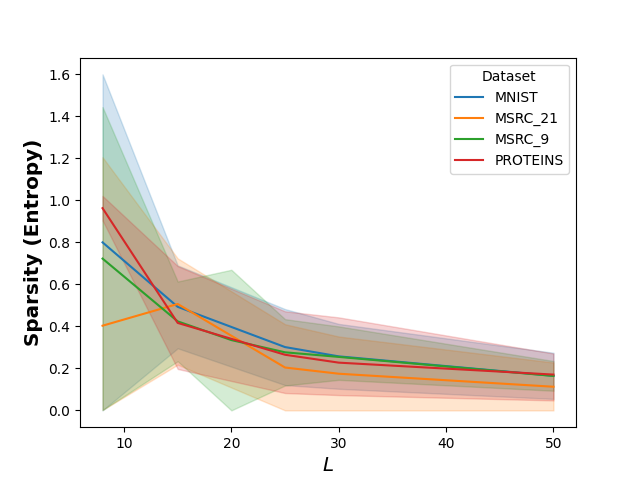}
		\caption{Von Neumann Entropy}
	\end{subfigure}
	\begin{subfigure}{0.3\textwidth}
		\includegraphics[width=\textwidth]{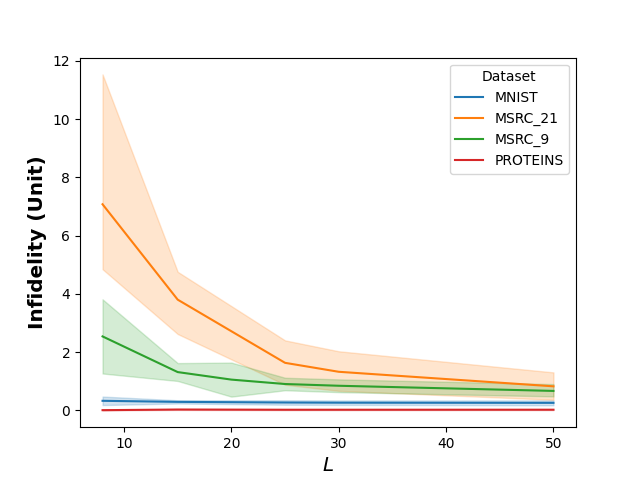}
		\caption{Infidelity (Unit)}
	\end{subfigure}
	\begin{subfigure}{0.3\textwidth}
		\includegraphics[width=\textwidth]{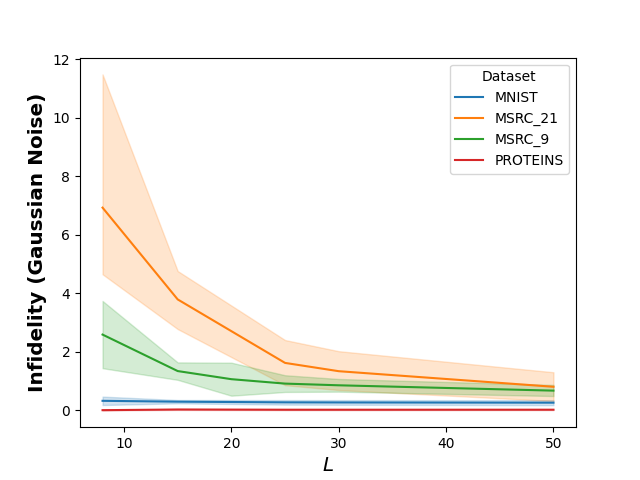}
		\caption{Infidelity (Gaussian noise)}
	\end{subfigure}
	\caption{Impact of $L$ regarding objective metrics}
	\label{L_bench}
\end{figure}
For the concept basis width, we recover numerically our statement regarding the fact that large argument basis favors to produce expressive explanations which are prone to be on one hand less infidel and, on the other hand, more concise as shown in Figure \ref{L_bench}.
\section{Conclusion}
It is common to encounter deep learning models and especially GNN for tackling academic and industrial problems notably in sensitive contexts such as healthcare, autonomous driving, etc.
Their powerfulness is often at the expense of having humanly unintelligible
decision processes that these models design. It raises serious issues for a safety
deployment of these models in our society.
We need to explain their inner working in order to gain insights and
rendering them trustworthy. Nonetheless, explaining processes are profitable only
and only if explanations are suited to the explainee.
State-of-the-art methods often provide absolute explanation regardless explainee
background or expectation and fail to include such explainee dependency although
widely discussed in the literature.
In this study, we address this concern with EiX-GNN, a new approach that fully
integrates this ubiquitous dependency with defining the explainee concept
assimibility notion allowing to adapt the explanation process to the explainee.
We lead a qualitative study in regards with this social aspect over real-world data and we compared, with respect to objective metrics used in literature, the fairness and compactness properties in comparison with relevant state-of-the-art methods.
In both settings, we provide meaningful results by, addressing the explainee-dependency issue and, outperforming state-of-the-art methods according to widely used objective metrics.


\clearpage
\bibliographystyle{plain}
\bibliography{biblio}

\begin{thebibliography}{10}

\bibitem{noauthor_definition_nodate}
Definition of {INTERPRET}.

\bibitem{baldassarre_explainability_2019}
Federico Baldassarre and Hossein Azizpour.
\newblock Explainability {Techniques} for {Graph} {Convolutional} {Networks},
  May 2019.
\newblock arXiv:1905.13686 [cs, stat].

\bibitem{bapst_unveiling_2020}
V.~Bapst, T.~Keck, A.~Grabska-Barwi{\'n}ska, C.~Donner, E.~D. Cubuk, S.~S.
  Schoenholz, A.~Obika, A.~W.~R. Nelson, T.~Back, D.~Hassabis, and P.~Kohli.
\newblock Unveiling the predictive power of static structure in glassy systems.
\newblock {\em Nat. Phys.}, 16(4):448--454, April 2020.
\newblock Number: 4 Publisher: Nature Publishing Group.

\bibitem{bas_scientific_1980}
C.~Van~Fraassen Bas.
\newblock {\em The {Scientific} {Image}}.
\newblock Oxford, England: Oxford University Press, 1980.

\bibitem{battaglia_relational_2018}
Peter~W. Battaglia, Jessica~B. Hamrick, Victor Bapst, Alvaro Sanchez-Gonzalez,
  Vinicius Zambaldi, Mateusz Malinowski, Andrea Tacchetti, David Raposo, Adam
  Santoro, Ryan Faulkner, Caglar Gulcehre, Francis Song, Andrew Ballard, Justin
  Gilmer, George Dahl, Ashish Vaswani, Kelsey Allen, Charles Nash, Victoria
  Langston, Chris Dyer, Nicolas Heess, Daan Wierstra, Pushmeet Kohli, Matt
  Botvinick, Oriol Vinyals, Yujia Li, and Razvan Pascanu.
\newblock Relational inductive biases, deep learning, and graph networks,
  October 2018.
\newblock arXiv:1806.01261 [cs, stat].

\bibitem{bechtel_explanation_2005}
William Bechtel and Adele Abrahamsen.
\newblock Explanation: a mechanist alternative.
\newblock {\em Studies in History and Philosophy of Science Part C: Studies in
  History and Philosophy of Biological and Biomedical Sciences},
  36(2):421--441, June 2005.

\bibitem{borgwardt_protein_2005}
Karsten~M. Borgwardt, Cheng~Soon Ong, Stefan Sch{\"o}nauer, S.~V.~N.
  Vishwanathan, Alex~J. Smola, and Hans-Peter Kriegel.
\newblock Protein function prediction via graph kernels.
\newblock {\em Bioinformatics}, 21(suppl\_1):i47--i56, June 2005.

\bibitem{bronstein_geometric_2021}
Michael~M. Bronstein, Joan Bruna, Taco Cohen, and Petar Veli{\v c}kovi{\'c}.
\newblock Geometric {Deep} {Learning}: {Grids}, {Groups}, {Graphs},
  {Geodesics}, and {Gauges}, May 2021.
\newblock arXiv:2104.13478 [cs, stat].

\bibitem{chater_mental_2006}
Nick Chater and Mike Oaksford.
\newblock Mental {Mechanisms}: {Speculations} on {Human} {Causal} {Learning}
  and {Reasoning}.
\newblock In {\em Information sampling and adaptive cognition}, pages 210--236.
  Cambridge University Press, New York, NY, US, 2006.

\bibitem{clough_global_2019}
James~R. Clough, Ilkay Oksuz, Esther Puyol-Ant{\'o}n, Bram Ruijsink, Andrew~P.
  King, and Julia~A. Schnabel.
\newblock Global and {Local} {Interpretability} for {Cardiac} {MRI}
  {Classification}.
\newblock In {\em Medical {Image} {Computing} and {Computer} {Assisted}
  {Intervention} {\textendash} {MICCAI} 2019: 22nd {International}
  {Conference}, {Shenzhen}, {China}, {October} 13{\textendash}17, 2019,
  {Proceedings}, {Part} {IV}}, pages 656--664, Berlin, Heidelberg, October
  2019. Springer-Verlag.

\bibitem{defferrard_convolutional_2016}
Micha{\"e}l Defferrard, Xavier Bresson, and Pierre Vandergheynst.
\newblock Convolutional {Neural} {Networks} on {Graphs} with {Fast} {Localized}
  {Spectral} {Filtering}.
\newblock In {\em Advances in {Neural} {Information} {Processing} {Systems}},
  volume~29. Curran Associates, Inc., 2016.

\bibitem{derrow-pinion_eta_2021}
Austin Derrow-Pinion, Jennifer She, David Wong, Oliver Lange, Todd Hester, Luis
  Perez, Marc Nunkesser, Seongjae Lee, Xueying Guo, Brett Wiltshire, Peter~W.
  Battaglia, Vishal Gupta, Ang Li, Zhongwen Xu, Alvaro Sanchez-Gonzalez, Yujia
  Li, and Petar Veli{\v c}kovi{\'c}.
\newblock {ETA} {Prediction} with {Graph} {Neural} {Networks} in {Google}
  {Maps}.
\newblock {\em Proceedings of the 30th ACM International Conference on
  Information \& Knowledge Management}, pages 3767--3776, October 2021.
\newblock arXiv: 2108.11482.

\bibitem{devlin_bert_2019}
Jacob Devlin, Ming-Wei Chang, Kenton Lee, and Kristina Toutanova.
\newblock {BERT}: {Pre}-training of {Deep} {Bidirectional} {Transformers} for
  {Language} {Understanding}.
\newblock {\em arXiv:1810.04805 [cs]}, May 2019.
\newblock arXiv: 1810.04805.

\bibitem{oliver_graphsvx_2021}
Alexandre Duval and Fragkiskos~D. Malliaros.
\newblock {GraphSVX}: {Shapley} {Value} {Explanations} for {Graph} {Neural}
  {Networks}.
\newblock In Nuria Oliver, Fernando P{\'e}rez-Cruz, Stefan Kramer, Jesse Read,
  and Jose~A. Lozano, editors, {\em Machine {Learning} and {Knowledge}
  {Discovery} in {Databases}. {Research} {Track}}, volume 12976, pages
  302--318. Springer International Publishing, Cham, 2021.
\newblock Series Title: Lecture Notes in Computer Science.

\bibitem{gilmer_neural_2017}
Justin Gilmer, Samuel~S. Schoenholz, Patrick~F. Riley, Oriol Vinyals, and
  George~E. Dahl.
\newblock Neural {Message} {Passing} for {Quantum} {Chemistry}.
\newblock In {\em Proceedings of the 34th {International} {Conference} on
  {Machine} {Learning}}, pages 1263--1272. PMLR, July 2017.
\newblock ISSN: 2640-3498.

\bibitem{glennan_rethinking_2002}
Stuart Glennan.
\newblock Rethinking {Mechanistic} {Explanation}.
\newblock {\em Philos. of Sci.}, 69(S3):S342--S353, September 2002.

\bibitem{gori_new_2005}
M.~Gori, G.~Monfardini, and F.~Scarselli.
\newblock A new model for learning in graph domains.
\newblock In {\em Proceedings. 2005 {IEEE} {International} {Joint} {Conference}
  on {Neural} {Networks}, 2005.}, volume~2, pages 729--734 vol. 2, July 2005.
\newblock ISSN: 2161-4407.

\bibitem{huang_graphlime_2022}
Qiang Huang, Makoto Yamada, Yuan Tian, Dinesh Singh, and Yi~Chang.
\newblock {GraphLIME}: {Local} {Interpretable} {Model} {Explanations} for
  {Graph} {Neural} {Networks}.
\newblock {\em IEEE Transactions on Knowledge and Data Engineering}, pages
  1--6, 2022.
\newblock Conference Name: IEEE Transactions on Knowledge and Data Engineering.

\bibitem{hurley_comparing_2009}
Niall Hurley and Scott Rickard.
\newblock Comparing {Measures} of {Sparsity}.
\newblock {\em IEEE Transactions on Information Theory}, 55(10):4723--4741,
  October 2009.
\newblock Conference Name: IEEE Transactions on Information Theory.

\bibitem{jacovi_towards_2020}
Alon Jacovi and Yoav Goldberg.
\newblock Towards {Faithfully} {Interpretable} {NLP} {Systems}: {How} {Should}
  {We} {Define} and {Evaluate} {Faithfulness}?
\newblock In {\em Proceedings of the 58th {Annual} {Meeting} of the
  {Association} for {Computational} {Linguistics}}, pages 4198--4205, Online,
  July 2020. Association for Computational Linguistics.

\bibitem{jacovi_aligning_2021}
Alon Jacovi and Yoav Goldberg.
\newblock Aligning {Faithful} {Interpretations} with their {Social}
  {Attribution}.
\newblock {\em Transactions of the Association for Computational Linguistics},
  9:294--310, March 2021.

\bibitem{keil_explanation_2006}
Frank~C. Keil.
\newblock Explanation and {Understanding}.
\newblock {\em Annu Rev Psychol}, 57:227--254, 2006.

\bibitem{kim_interpretability_nodate}
Been Kim, Martin Wattenberg, Justin Gilmer, Carrie Cai, James Wexler, Fernanda
  Viegas, and Rory Sayres.
\newblock Interpretability {Beyond} {Feature} {Attribution}: {Quantitative}
  {Testing} with {Concept} {Activation} {Vectors} ({TCAV}).
\newblock page~10.

\bibitem{kingma_adam_2017}
Diederik~P. Kingma and Jimmy Ba.
\newblock Adam: {A} {Method} for {Stochastic} {Optimization}.
\newblock Technical Report arXiv:1412.6980, arXiv, January 2017.
\newblock arXiv:1412.6980 [cs] type: article.

\bibitem{kipf_semi-supervised_2017}
Thomas~N. Kipf and Max Welling.
\newblock Semi-{Supervised} {Classification} with {Graph} {Convolutional}
  {Networks}.
\newblock {\em arXiv:1609.02907 [cs, stat]}, February 2017.
\newblock arXiv: 1609.02907.

\bibitem{krizhevsky_imagenet_2012}
Alex Krizhevsky, Ilya Sutskever, and Geoffrey~E Hinton.
\newblock {ImageNet} {Classification} with {Deep} {Convolutional} {Neural}
  {Networks}.
\newblock In {\em Advances in {Neural} {Information} {Processing} {Systems}},
  volume~25. Curran Associates, Inc., 2012.

\bibitem{kulesza_too_2013}
Todd Kulesza, Simone Stumpf, Margaret Burnett, Sherry Yang, Irwin Kwan, and
  Weng-Keen Wong.
\newblock Too much, too little, or just right? {Ways} explanations impact end
  users' mental models.
\newblock In {\em 2013 {IEEE} {Symposium} on {Visual} {Languages} and {Human}
  {Centric} {Computing}}, pages 3--10, September 2013.
\newblock ISSN: 1943-6106.

\bibitem{lecun_gradient-based_1998}
Y.~Lecun, L.~Bottou, Y.~Bengio, and P.~Haffner.
\newblock Gradient-based learning applied to document recognition.
\newblock {\em Proceedings of the IEEE}, 86(11):2278--2324, November 1998.
\newblock Conference Name: Proceedings of the IEEE.

\bibitem{lipton_mythos_2018}
Zachary~C. Lipton.
\newblock The {Mythos} of {Model} {Interpretability}: {In} machine learning,
  the concept of interpretability is both important and slippery.
\newblock {\em Queue}, 16(3):31--57, June 2018.

\bibitem{lucic_cf-gnnexplainer_2022}
Ana Lucic, Maartje A.~Ter Hoeve, Gabriele Tolomei, Maarten~De Rijke, and
  Fabrizio Silvestri.
\newblock {CF}-{GNNExplainer}: {Counterfactual} {Explanations} for {Graph}
  {Neural} {Networks}.
\newblock In {\em Proceedings of {The} 25th {International} {Conference} on
  {Artificial} {Intelligence} and {Statistics}}, pages 4499--4511. PMLR, May
  2022.
\newblock ISSN: 2640-3498.

\bibitem{luo_parameterized_2020}
Dongsheng Luo, Wei Cheng, Dongkuan Xu, Wenchao Yu, Bo~Zong, Haifeng Chen, and
  Xiang Zhang.
\newblock Parameterized {Explainer} for {Graph} {Neural} {Network}.
\newblock In {\em Advances in {Neural} {Information} {Processing} {Systems}},
  volume~33, pages 19620--19631. Curran Associates, Inc., 2020.

\bibitem{miller_explanation_2019}
Tim Miller.
\newblock Explanation in artificial intelligence: {Insights} from the social
  sciences.
\newblock {\em Artificial Intelligence}, 267:1--38, February 2019.

\bibitem{monti_geometric_2017}
Federico Monti, Davide Boscaini, Jonathan Masci, Emanuele Rodola, Jan Svoboda,
  and Michael~M. Bronstein.
\newblock Geometric {Deep} {Learning} on {Graphs} and {Manifolds} {Using}
  {Mixture} {Model} {CNNs}.
\newblock pages 5115--5124, 2017.

\bibitem{monti_dual-primal_2018}
Federico Monti, Oleksandr Shchur, Aleksandar Bojchevski, Or~Litany, Stephan
  G{\"u}nnemann, and Michael~M. Bronstein.
\newblock Dual-{Primal} {Graph} {Convolutional} {Networks}, June 2018.
\newblock arXiv:1806.00770 [cs, stat].

\bibitem{murdoch_definitions_2019}
W.~James Murdoch, Chandan Singh, Karl Kumbier, Reza Abbasi-Asl, and Bin Yu.
\newblock Definitions, methods, and applications in interpretable machine
  learning.
\newblock {\em Proceedings of the National Academy of Sciences},
  116(44):22071--22080, October 2019.
\newblock Publisher: Proceedings of the National Academy of Sciences.

\bibitem{page_pagerank_1999}
Lawrence Page, Sergey Brin, Rajeev Motwani, and Terry Winograd.
\newblock The {PageRank} {Citation} {Ranking}: {Bringing} {Order} to the
  {Web}., November 1999.
\newblock Publisher: Stanford InfoLab.

\bibitem{pope_explainability_2019}
Phillip~E. Pope, Soheil Kolouri, Mohammad Rostami, Charles~E. Martin, and Heiko
  Hoffmann.
\newblock Explainability {Methods} for {Graph} {Convolutional} {Neural}
  {Networks}.
\newblock In {\em 2019 {IEEE}/{CVF} {Conference} on {Computer} {Vision} and
  {Pattern} {Recognition} ({CVPR})}, pages 10764--10773, Long Beach, CA, USA,
  June 2019. IEEE.

\bibitem{russakovsky_imagenet_2015}
Olga Russakovsky, Jia Deng, Hao Su, Jonathan Krause, Sanjeev Satheesh, Sean Ma,
  Zhiheng Huang, Andrej Karpathy, Aditya Khosla, Michael Bernstein,
  Alexander~C. Berg, and Li~Fei-Fei.
\newblock {ImageNet} {Large} {Scale} {Visual} {Recognition} {Challenge}.
\newblock {\em Int J Comput Vis}, 115(3):211--252, December 2015.

\bibitem{sanchez-lengeling_evaluating_2020}
Benjamin Sanchez-Lengeling, Jennifer Wei, Brian Lee, Emily Reif, Peter Wang,
  Wesley Qian, Kevin McCloskey, Lucy Colwell, and Alexander Wiltschko.
\newblock Evaluating {Attribution} for {Graph} {Neural} {Networks}.
\newblock In {\em Advances in {Neural} {Information} {Processing} {Systems}},
  volume~33, pages 5898--5910. Curran Associates, Inc., 2020.

\bibitem{satorras_en_2021}
V{\i}?ctor~Garcia Satorras, Emiel Hoogeboom, and Max Welling.
\newblock E(n) {Equivariant} {Graph} {Neural} {Networks}.
\newblock In {\em Proceedings of the 38th {International} {Conference} on
  {Machine} {Learning}}, pages 9323--9332. PMLR, July 2021.
\newblock ISSN: 2640-3498.

\bibitem{scarselli_graph_2009}
Franco Scarselli, Marco Gori, Ah~Chung Tsoi, Markus Hagenbuchner, and Gabriele
  Monfardini.
\newblock The {Graph} {Neural} {Network} {Model}.
\newblock {\em IEEE Transactions on Neural Networks}, 20(1):61--80, January
  2009.
\newblock Conference Name: IEEE Transactions on Neural Networks.

\bibitem{schlichtkrull_incorporating_2021}
Michael~Sejr Schlichtkrull.
\newblock {\em Incorporating {Structure} into {Neural} {Models} for {Language}
  {Processing}}.
\newblock doctoral, University of Amsterdam, May 2021.

\bibitem{schnake_higher-order_2021}
Thomas Schnake, Oliver Eberle, Jonas Lederer, Shinichi Nakajima, Kristof~T.
  Sch{\"u}tt, Klaus-Robert M{\"u}ller, and Gr{\'e}goire Montavon.
\newblock Higher-{Order} {Explanations} of {Graph} {Neural} {Networks} via
  {Relevant} {Walks}.
\newblock {\em IEEE Trans. Pattern Anal. Mach. Intell.}, pages 1--1, 2021.
\newblock arXiv:2006.03589 [cs, stat].

\bibitem{selbst_fairness_2019}
Andrew~D. Selbst, Danah Boyd, Sorelle~A. Friedler, Suresh Venkatasubramanian,
  and Janet Vertesi.
\newblock Fairness and {Abstraction} in {Sociotechnical} {Systems}.
\newblock In {\em Proceedings of the {Conference} on {Fairness},
  {Accountability}, and {Transparency}}, {FAT}* '19, pages 59--68, New York,
  NY, USA, January 2019. Association for Computing Machinery.

\bibitem{shapley_notes_1951}
Lloyd~S. Shapley.
\newblock Notes on the {N}-{Person} {Game} {\textemdash} {II}: {The} {Value} of
  an {N}-{Person} {Game}.
\newblock Technical report, RAND Corporation, August 1951.

\bibitem{shotton_textonboost_2009}
Jamie Shotton, John Winn, Carsten Rother, and Antonio Criminisi.
\newblock {TextonBoost} for {Image} {Understanding}: {Multi}-{Class} {Object}
  {Recognition} and {Segmentation} by {Jointly} {Modeling} {Texture}, {Layout},
  and {Context}.
\newblock {\em Int J Comput Vis}, 81(1):2--23, January 2009.

\bibitem{vaswani_attention_nodate}
Ashish Vaswani, Noam Shazeer, Niki Parmar, Jakob Uszkoreit, Llion Jones,
  Aidan~N Gomez, {\L }ukasz Kaiser, and Illia Polosukhin.
\newblock Attention is {All} you {Need}.
\newblock page~11.

\bibitem{velickovic_graph_2018}
Petar Veli{\v c}kovi{\'c}, Guillem Cucurull, Arantxa Casanova, Adriana Romero,
  Pietro Li{\`o}, and Yoshua Bengio.
\newblock Graph {Attention} {Networks}.
\newblock {\em arXiv:1710.10903 [cs, stat]}, February 2018.
\newblock arXiv: 1710.10903.

\bibitem{vu_pgm-explainer_2020}
Minh Vu and My~T. Thai.
\newblock {PGM}-{Explainer}: {Probabilistic} {Graphical} {Model} {Explanations}
  for {Graph} {Neural} {Networks}.
\newblock In {\em Advances in {Neural} {Information} {Processing} {Systems}},
  volume~33, pages 12225--12235. Curran Associates, Inc., 2020.

\bibitem{winn_object_2005}
J.~Winn, A.~Criminisi, and T.~Minka.
\newblock Object categorization by learned universal visual dictionary.
\newblock In {\em Tenth {IEEE} {International} {Conference} on {Computer}
  {Vision} ({ICCV}'05) {Volume} 1}, volume~2, pages 1800--1807 Vol. 2, October
  2005.
\newblock ISSN: 2380-7504.

\bibitem{yanardag_deep_2015}
Pinar Yanardag and S.V.N. Vishwanathan.
\newblock Deep {Graph} {Kernels}.
\newblock In {\em Proceedings of the 21th {ACM} {SIGKDD} {International}
  {Conference} on {Knowledge} {Discovery} and {Data} {Mining}}, {KDD} '15,
  pages 1365--1374, New York, NY, USA, August 2015. Association for Computing
  Machinery.

\bibitem{yeh_fidelity_2019}
Chih-Kuan Yeh, Cheng-Yu Hsieh, Arun Suggala, David~I Inouye, and Pradeep~K
  Ravikumar.
\newblock On the ({In})fidelity and {Sensitivity} of {Explanations}.
\newblock In {\em Advances in {Neural} {Information} {Processing} {Systems}},
  volume~32. Curran Associates, Inc., 2019.

\bibitem{ying_graph_2018}
Rex Ying, Ruining He, Kaifeng Chen, Pong Eksombatchai, William~L. Hamilton, and
  Jure Leskovec.
\newblock Graph {Convolutional} {Neural} {Networks} for {Web}-{Scale}
  {Recommender} {Systems}.
\newblock {\em Proceedings of the 24th ACM SIGKDD International Conference on
  Knowledge Discovery \& Data Mining}, pages 974--983, July 2018.
\newblock arXiv: 1806.01973.

\bibitem{ying_gnnexplainer_2019}
Zhitao Ying, Dylan Bourgeois, Jiaxuan You, Marinka Zitnik, and Jure Leskovec.
\newblock {GNNExplainer}: {Generating} {Explanations} for {Graph} {Neural}
  {Networks}.
\newblock In {\em Advances in {Neural} {Information} {Processing} {Systems}},
  volume~32. Curran Associates, Inc., 2019.

\bibitem{yuan_xgnn_2020}
Hao Yuan, Jiliang Tang, Xia Hu, and Shuiwang Ji.
\newblock {XGNN}: {Towards} {Model}-{Level} {Explanations} of {Graph} {Neural}
  {Networks}.
\newblock In {\em Proceedings of the 26th {ACM} {SIGKDD} {International}
  {Conference} on {Knowledge} {Discovery} \& {Data} {Mining}}, pages 430--438.
  Association for Computing Machinery, New York, NY, USA, August 2020.

\bibitem{yuan_explainability_2021}
Hao Yuan, Haiyang Yu, Jie Wang, Kang Li, and Shuiwang Ji.
\newblock On {Explainability} of {Graph} {Neural} {Networks} via {Subgraph}
  {Explorations}.
\newblock In {\em Proceedings of the 38th {International} {Conference} on
  {Machine} {Learning}}, pages 12241--12252. PMLR, July 2021.
\newblock ISSN: 2640-3498.

\bibitem{zhang_link_2018}
Muhan Zhang and Yixin Chen.
\newblock Link {Prediction} {Based} on {Graph} {Neural} {Networks}.
\newblock In {\em Advances in {Neural} {Information} {Processing} {Systems}},
  volume~31. Curran Associates, Inc., 2018.

\end{thebibliography}
\clearpage
\appendix
\section{Experimental setup details}

\subsection{Datasets details}
\label{data:dataset}
In order to provide meaningful results, we chose real-world datasets that incorporate human intelligible features. These datasets are often used \cite{baldassarre_explainability_2019,oliver_graphsvx_2021,huang_graphlime_2022,lucic_cf-gnnexplainer_2022,pope_explainability_2019,schnake_higher-order_2021,ying_gnnexplainer_2019,yuan_xgnn_2020,yuan_explainability_2021} to illustrate explaining methods for GNN-based classifiers. Each of the following datasets is suited for graph classification problems.
Note that REDDIT-BINARY does not need any prior knowledge to assess any explanation map and groundtruth explanation is easy to consider as mentioned in \cite{ying_gnnexplainer_2019}, on the contrary to PROTEINS which require chemical knowledge.
In the following, we found details regarding used datasets:
\paragraph{MNISTSuperpixel}
\cite{monti_geometric_2017} is a dataset composed of 60000 graphs, that each represents a superpixel version of the well-known handwritten digit MNIST \cite{lecun_gradient-based_1998} dataset. Each MNISTSuperpixels instance is a graph representation of the original MNIST instance. Two vertices are linked according to their spatial proximity.
\paragraph{PROTEINS}
\cite{borgwardt_protein_2005} is a dataset counting 1113 labeled graphs. Each graph represents a protein that is classified as enzymes or non-enzymes. Nodes represent the amino acids and two nodes are connected if they also share the same spatial locality.
\paragraph{MSRC}
\cite{shotton_textonboost_2009,winn_object_2005} datasets are used in image semantic segmentation problems. Each image in converted into a semantic superpixel version of it. In MSRC-9, which is composed of 221 labeled graphs, semantics label are distributed among 8 semantic labels. In the MSRC-21 version, composed of 563 labeled graphs, extends the number of possible semantic labels to 21.
\paragraph{REDDIT-BINARY}
\cite{yanardag_deep_2015} is a dataset composed of 2000 graphs where each of them represents a question/answer-based thread of Reddit, namely \textit{r/IAmA} and \textit{r/AskReddit}. In these graphs, node represents users and there is a link between two users if one has answered the other.

\subsection{Classifier details}
\label{data:classifier}
Here the general framework of graph classification problems under the view of \gls{gnn} models.
\subsubsection{Supervised graph classification problems}
For $\mathcal{G},\mathcal{Y}$ two measurable spaces, we define $\mathcal{F}(\mathcal{G},\mathcal{Y})$ the set of measurable functions going from $\mathcal{G}$ to $\mathcal{Y}$.
Given an i.i.d sampled finite dataset $\mathcal{D}\subset\mathcal{G}\times\mathcal{Y}$ where each element $Z_i=(G_i,Y_i)$ is a graph $G_i$ and its label $Y_i$ representing the class it belongs to.
A loss function is mapping
$\mathcal{L}: \mathcal{F}(\mathcal{G},\mathcal{Y})\times\mathcal{D}\rightarrow\mathbb{R}$
quantifying how well a learning mapping
$f\in\mathcal{F}_{a,\theta}\subset\mathcal{F}(\mathcal{G},\mathcal{Y})$
associated $G_i$ to its true label $Y_i$ conditioned by a neural network architecture $a$ and
a learning parameter $\theta$. For a given architecture $\hat{a}$, we seek $f^{\star}$ such that:
\begin{equation}
	f^{\star}=\underset{f_{\theta}\in\mathcal{F}_{\hat{a},\theta}}{\arg\min}\underset{Z\sim\hat{\mathcal{D}}}{\mathbb{E}}\left[\mathcal{L}(f_{\theta},Z)\right] \text{\hspace{0.2cm} with \hspace{0.2cm}} \underset{Z\sim\hat{\mathcal{D}}}{\mathbb{E}}\left[\mathcal{L}(f_{\theta},Z)\right]=\int_{\hat{\mathcal{D}}}\mathcal{L}(f_{\theta},z)d\mathbb{P}_{Z}(z)
\end{equation}
where $\hat{\mathcal{D}}$ is $f_{\theta}$-unseen data and $Z=(G,Y)$ where $G$ is a $\mathcal{G}$-valued random variable, $Y$ is a $\mathcal{Y}$-valued random variable and $\mathbb{P}_Z$ is the image probability measure of $Z$ in $\hat{\mathcal{D}}$.
In the context of graph classification, $f$ is a \gls{gnn} model and
$\mathcal{L}$ is the cross-entropy loss between the inferred label conditional probability law and its ground truth-conditional probability law.

\subsubsection{Learning details}
Except for the classification task on MNISTSuperpixel, we have trained two \gls{gnn} models :  one based on \gls{gcn} \cite{kipf_semi-supervised_2017} and the other based on GAT \cite{velickovic_graph_2018} that we name here generically as the descriptor module of the classifier.
We chained two descriptor modules then we feed outputs to a global average
pooling layer, a linear module is then used to classify with softmax function.
In between layers, we use Relu function as activation function.  For the MNISTSuperpixels dataset, we use four chained descriptor modules and $\tanh$ as an activation function.
All these different implementations use the ADAM \cite{kingma_adam_2017} version of the stochastic gradient descend approach with the same learning parameter equals to $10^{-4}$. We use an Intel $\copyright$ Xeon Silver 4208 and Nvidia $\copyright$ Tesla A100 40 GB GPU for our training during 100 epochs.
Under this consideration, we have obtained for each dataset an accurate classifier as accurate as those used in comparing method experiments.

\subsection{Comparing method details}
\label{data:comparing}
As comparing methods, we have used three black-box model-based local post-hoc methods that have achieved strong results in the literature. We give here additional details regarding these methods.
\paragraph{GNNExplainer} \cite{ying_gnnexplainer_2019}  is a local post-hoc
model-based explaining method suited for GNN. It looks after which subgraphs, derived
from the input graph, contain the highest mutual information with
the this one. This method has achieved strong results on many explanations problems.
\paragraph{SubgraphX} \cite{yuan_explainability_2021} is also a local post-hoc
model-based explaining method suited for GNN. From the input graph, it uses a
Monte Carlo Tree Search method to find heuristically relevant substructures for explanations purposes.
\paragraph{PGExplainer} \cite{luo_parameterized_2020} share the same idea as
GNNExplainer but is rather concentrated on which edges in the input graph is
important to conserve the classifier expressivity.
\section{Objective assessment metric details}
Explaining internal decision processes involved in classification problems are often linked with the necessity to assess obtained saliency maps meaningfulness. Thus it requires task-related expert assessment which is subjective and consequently biased. Literature has proposed several objective metrics that are expert independent to evaluate quantitatively the quality of explanation maps.
\paragraph{Infidelity}\cite{yeh_fidelity_2019} quantifies in which manner the explanation maps provided by an explanation mapping $\phi$ of predictions made by an optimized classifier $f^{\star}$ change when an input $\X$ is perturbed by a random variable $\textbf{I}$ following a perturbation density $B$. It is defined by:
\begin{equation}
	\texttt{Infd}(\phi,f^{\star},\X)=\underset{\textbf{I}\sim B}{\mathbb{E}}[(\textbf{I}^{T}\boldsymbol{\phi}(\X,f^{\star})-(f^{\star}(\X)-f^{\star}(\X-\textbf{I})))^2]
\end{equation}
The perturbing distribution $B$ is used to be a standard normal distribution, as mentioned in \cite{yeh_fidelity_2019}.
\paragraph{Sparsity} Generally speaking, concise explanations are preferred than wide explanations that drown pertinent information.
This statement does not dependent on the context of the explanations, so it is an objective statement. The Von Neumann entropy appears to be a good candidate for measuring such sparsity \cite{hurley_comparing_2009} . The Von Neumann entropy of a probability distribution encodes the uncertainty amount induced in this probability distribution. It can be seen as a sparsity metric since if the distribution mass is spatially concentrated on the domain (i.e., lower entropy) it induced that explanation arguments are clearly identified.
On the contrary, if the entropy is important, it means that explicative elements are blurry diffused and scattered on the domain which is less insightful for the user. For a probability distribution $\boldsymbol{\pi}\in[0,1]^d$ it is defined as :
\begin{equation}
	H(\boldsymbol{\pi})=-\underset{\boldsymbol{\pi}}{\mathbb{E}}\left[\ln(\boldsymbol{\pi})\right]
\end{equation}

\subsection{Summarized quantitative results}
\label{data:objmet}
Here we have summarized quantitive results we have obtained when we have benchmarked our methods with comparing methods over real-world datasets. As mentioned by the $\downarrow$ symbol, the lower the better.

\begin{table}[H]
	\centering
	\begin{tabular}{@{}ccccc@{}}
		\textbf{Dataset}                                                                               &
		\textbf{Explainer}                                                                             &
		\textbf{Entropy}  $(\downarrow)$                                                               &
		\textbf{Infidelity (Gaussian)} $(\downarrow)$                                                  &
		\textbf{Infidelity (Unit)} $(\downarrow)$                                                        \\
		\hline
		\multicolumn{1}{l}{}                                                                           &
		\multicolumn{1}{l}{}                                                                           &
		\multicolumn{1}{l}{}                                                                           &
		\multicolumn{1}{l}{}                                                                           &
		\multicolumn{1}{l}{}                                                                             \\
		MNISTSuperpixels                                                                               &
		\begin{tabular}[c]{@{}c@{}}EiX-GNN\\ GNNExplainer \\ PGExplainer \\ SubgraphX \end{tabular}    &
		\begin{tabular}[c]{@{}c@{}}\textbf{9.41E-01} \\1.30E+03 \\ 1.21E+03 \\ 1.70E+02\end{tabular}   &
		\begin{tabular}[c]{@{}c@{}}\textbf{5.69E+00} \\2.43E+05 \\ 1.80E+04 \\ 1.31E+03  \end{tabular} &
		\begin{tabular}[c]{@{}c@{}}\textbf{5.69E+00} \\2.44E+05 \\ 1.80E+04 \\ 1.31E+04 \end{tabular}    \\
		\multicolumn{1}{l}{}                                                                           &
		\multicolumn{1}{l}{}                                                                           &
		\multicolumn{1}{l}{}                                                                           &
		\multicolumn{1}{l}{}                                                                           &
		\multicolumn{1}{l}{}                                                                             \\
		PROTEINS                                                                                       &
		\begin{tabular}[c]{@{}c@{}}EiX-GNN\\ GNNExplainer \\ PGExplainer \\ SubgraphX \end{tabular}    &
		\begin{tabular}[c]{@{}c@{}}\textbf{9.37E-01} \\ 5.21E+01 \\ 7.02E+01 \\ 1.40E+01 \end{tabular} &
		\begin{tabular}[c]{@{}c@{}}\textbf{2.38E-01} \\ 3.36E+02 \\ 8.23E+00 \\ 4.56E+01 \end{tabular} &
		\begin{tabular}[c]{@{}c@{}}\textbf{2.78E-01} \\ 3.47E+02 \\ 8.21E+00 \\ 4.56E+01 \end{tabular}   \\
		\multicolumn{1}{l}{}                                                                           &
		\multicolumn{1}{l}{}                                                                           &
		\multicolumn{1}{l}{}                                                                           &
		\multicolumn{1}{l}{}                                                                           &
		\multicolumn{1}{l}{}                                                                             \\
		MSRC-9                                                                                         &
		\begin{tabular}[c]{@{}c@{}}EiX-GNN\\ GNNExplainer \\ PGExplainer \\ SubgraphX \end{tabular}    &
		\begin{tabular}[c]{@{}c@{}}\textbf{9.02E-01} \\7.84E+01 \\ 8.69E+01 \\ 4.45E+01 \end{tabular}  &
		\begin{tabular}[c]{@{}c@{}}\textbf{2.29E-05} \\1.14E+03 \\ 2.31E+03 \\ 2.69E+03 \end{tabular}  &
		\begin{tabular}[c]{@{}c@{}}\textbf{9.68E-05} \\1.12E+03 \\ 2.29E+03 \\ 2.70E+03 \end{tabular}    \\
		\multicolumn{1}{l}{}                                                                           &
		\multicolumn{1}{l}{}                                                                           &
		\multicolumn{1}{l}{}                                                                           &
		\multicolumn{1}{l}{}                                                                           &
		\multicolumn{1}{l}{}                                                                             \\
		REDDIT-BINARY                                                                                  &
		\begin{tabular}[c]{@{}c@{}}EiX-GNN\\ GNNExplainer \\ PGExplainer \\ SubgraphX \end{tabular}    &
		\begin{tabular}[c]{@{}c@{}}\textbf{4.02E-01} \\ 6.71E+01 \\ 5.61E+01 \\ 3.95E+01 \end{tabular} &
		\begin{tabular}[c]{@{}c@{}}\textbf{2.64E-02} \\ 5.17E+03 \\ 2.35E+02 \\ 1.11E+03 \end{tabular} &
		\begin{tabular}[c]{@{}c@{}}\textbf{4.63E-01} \\ 5.14E+03 \\ 2.36E+02 \\ 1.110E+03 \end{tabular}  \\
		\multicolumn{1}{l}{}                                                                           &
		\multicolumn{1}{l}{}                                                                           &
		\multicolumn{1}{l}{}                                                                           &
		\multicolumn{1}{l}{}                                                                           &
		\multicolumn{1}{l}{}                                                                             \\
		MSRC-21                                                                                        &
		\begin{tabular}[c]{@{}c@{}}EiX-GNN\\ GNNExplainer \\ PGExplainer \\ SubgraphX \end{tabular}    &
		\begin{tabular}[c]{@{}c@{}}\textbf{8.54E-01} \\2.79E+02 \\ 1.90E+05 \\ 4.82E+03\end{tabular}   &
		\begin{tabular}[c]{@{}c@{}}\textbf{2.02E+00} \\1.03E+04 \\ 8.74E+02 \\ 3.67E+04\end{tabular}   &
		\begin{tabular}[c]{@{}c@{}}\textbf{2.05E+00} \\1.04E+04 \\ 8.74E+02 \\ 3.67E+04\end{tabular}     \\
	\end{tabular}
	\caption{Comparison between EiX-GNN and compared method over three objective quality assessment measures for benchmarked datasets}
	\label{tab:1}
\end{table}

\end{document}